%% file: main.tex
\documentclass[10pt,twocolumn,letterpaper]{article}

\usepackage{iccv}
\usepackage{times}
\usepackage{epsfig}
\usepackage{graphicx}
\usepackage{amsmath}
\usepackage{amssymb}
\usepackage{textcomp}
\usepackage{booktabs}
\usepackage{comment}
\usepackage{xcolor}
\usepackage{array}
\usepackage{multirow}
\usepackage{tabularx,adjustbox,booktabs}
\usepackage{soul}
\newcommand\setrow[1]{\gdef\rowmac{#1}#1\ignorespaces}
\newcommand\clearrow{\global\let\rowmac\relax}
\clearrow
\newcommand{\mohamed}{\textcolor{teal}}
\newcommand{\afsaneh}{\textcolor{red}}
\newcommand{\isma}{\textcolor{orange}}

\newcommand{\single}{\textit{(S) }}
\newcommand{\multiple}{\textit{(M) }}

\usepackage[pagebackref=true,breaklinks=true,letterpaper=true,colorlinks,bookmarks=false]{hyperref}

\def\iccvPaperID{9036} 

\iccvfinalcopy 
\ificcvfinal\fi

\newcommand{\PreserveBackslash}[1]{\let\temp=\\#1\let\\=\temp}
\newcolumntype{C}[1]{>{\PreserveBackslash\centering}p{#1}}
\newcolumntype{?}{!{\vrule width 1pt}}

\begin{document}

\title{\textsc{GePSAn}: \underline{Ge}nerative \underline{P}rocedure \underline{S}tep \underline{An}ticipation in Cooking Videos}

\author{Mohamed A. Abdelslam\textsuperscript{1}, Samrudhdhi B. Rangrej\textsuperscript{1*}, Isma Hadji\textsuperscript{1*}, Nikita Dvornik\textsuperscript{2*}, \\ Konstantinos G. Derpanis\textsuperscript{1,3}, Afsaneh Fazly\textsuperscript{1} \\
\textsuperscript{1}Samsung AI Centre\textbf{\ ,\ }\textsuperscript{2}Waabi\textbf{\ ,\ }\textsuperscript{3}York University \\
\texttt{\{m.abdelsalam, s.rangrej, isma.hadji, a.fazly\}@samsung.com}\textbf{\ ,\ } \\ \texttt{dvornik.nikita@gmail.com}\textbf{\ ,\ } \texttt{kosta@yorku.ca} \\\
}
\maketitle
\ificcvfinal\thispagestyle{empty}\fi

\begin{abstract}
We study the problem of future step anticipation in procedural videos. 
Given a video of an ongoing procedural activity, we predict a plausible next procedure step described in rich natural language.
While most previous work focuses on the problem of data scarcity in procedural video datasets, another core challenge of future anticipation is how to account for multiple plausible future realizations in natural settings. This problem has been largely overlooked in previous work.
To address this challenge, we frame future step prediction as modelling the distribution of all possible candidates for the next step. Specifically, we design a generative model that takes a series of video clips as input, and generates multiple plausible and diverse candidates (in natural language) for the next step.
Following previous work, we side-step the video annotation scarcity by pretraining our model on a large text-based corpus of procedural activities, and then transfer the model to the video domain.
Our experiments, both in textual and video domains, show that our model captures diversity in the next step prediction and generates multiple plausible future predictions. 
Moreover, our model establishes new state-of-the-art results on YouCookII, where it outperforms existing baselines on the next step anticipation.
Finally, we also show that our model can successfully transfer from text to the video domain zero-shot, i.e., without fine-tuning or adaptation, and produces good-quality future step predictions from video.
\end{abstract}

{\let\thefootnote\relax\footnotetext{*Equal Contribution}}
\section{Introduction}
\label{sec:intro}
\input{./Sections/introduction}

\section{Related work}
\label{sec:related_work}
\input{./Sections/related_work}

\section{Technical approach}
\label{sec:approach}
\input{./Sections/approach}

\section{Experiments}
\label{sec:experiments}
\input{./Sections/experiments}
\section{Conclusion}
\label{sec:conclusion}
\input{./Sections/conclusion}

{\small
\bibliographystyle{ieee_fullname}
\bibliography{main}
}
\clearpage
\onecolumn
\section{Supplementary}
\label{sec:appendix}
\input{./Sections/supplementary}
\end{document}

%% file: Sections/introduction.tex
Anticipating future steps while performing a task is a natural human behaviour necessary to successfully accomplish a task and cooperate with other humans.
Thus, it is important for a smart AI agent to exhibit this behaviour too, in order to assist humans in performing procedural tasks (\eg, cooking, assembling furniture or setting up an electronic device).
For example, consider a cooking AI assistant that observes a user as they cook a dish. To be useful, this assistant needs to anticipate possible next steps in order to provide timely support with ingredients, cooking tools and actions.
Anticipating future steps from a video stream is a challenging task, where simply recognizing the current action or objects is not sufficient.
To anticipate future actions, one needs to parse and recognize the human-object interactions from an unstructured and cluttered environment in the current frame, predict the possible task being performed (possibly leveraging the past observations) and finally anticipate plausible next steps.
Given the importance and challenges associated with this task, several research efforts targeted this application in the recent years \cite{sener.etal.2022,sener.yao.2019,girdhar2021anticipative,zhao.wildes.2020}.

\begin{figure}[t]
\centering
\includegraphics[trim=5 15 10 30,clip,width=0.99\columnwidth]{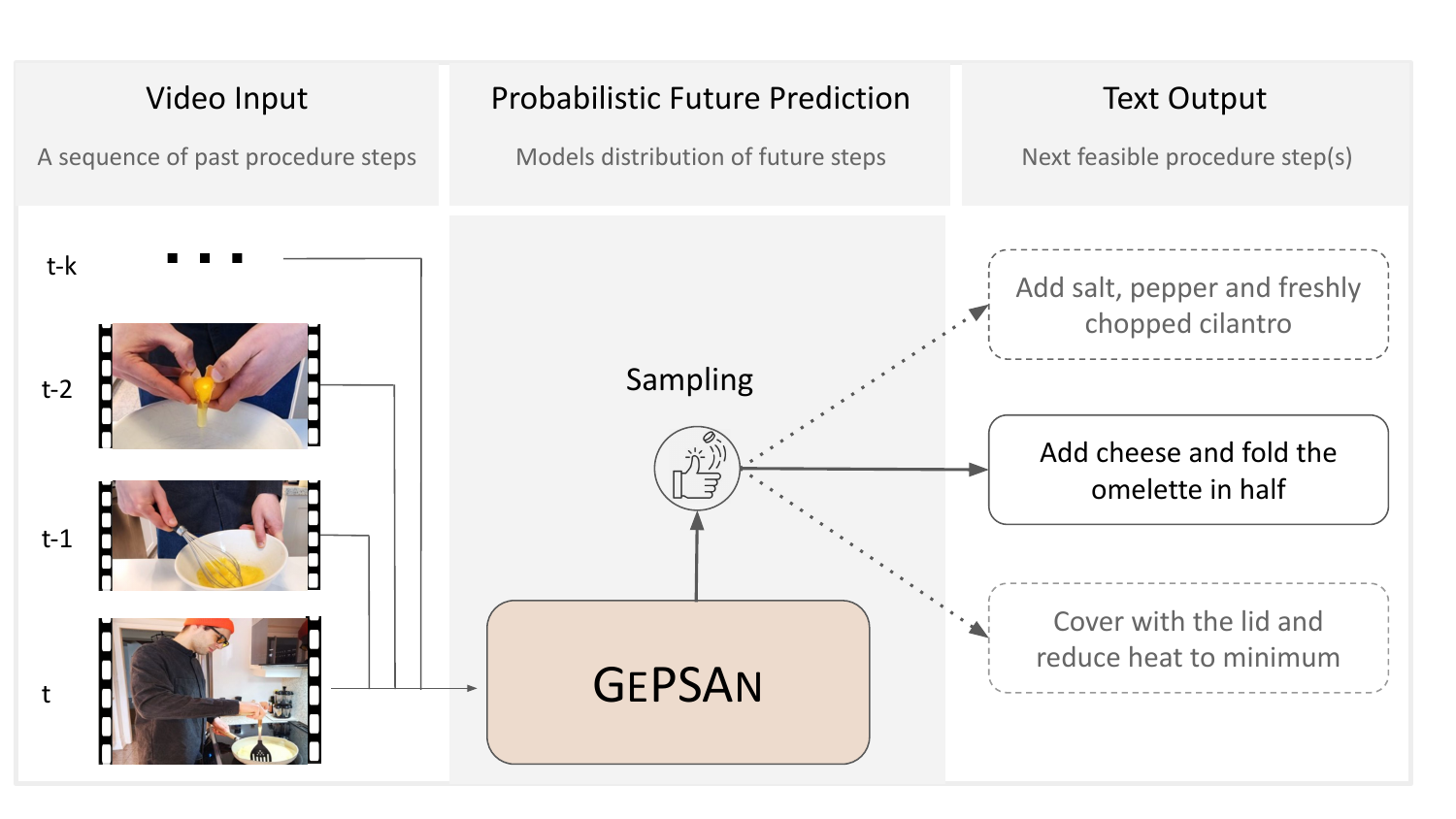}
\vspace{5pt}
\caption{\textbf{Summary of the proposed \textsc{GePSAn} model.}
Our model, given an initial video stream representing a sequence of past procedural steps, predicts multiple feasible alternatives for the next step in natural language. We first train our model on text-only data, followed by zero-shot transfer to the video domain.
\label{fig:teaser}
}
\end{figure}

We follow recent work \cite{sener.etal.2022} and tackle future anticipation in the realm of cooking activities. Given visual observations of the first $t$ steps, our task is to predict the next step to be executed.
This task entails recognizing the current step and the recipe being made, which is particularly challenging given the modest size of cooking video datasets with annotations.
Fortunately, such instructional knowledge is available in abundance in the text domain (think of all the dish recipes online) and can be leveraged to help video prediction.  
 Prior work \cite{sener.etal.2022,yang.etal.2022} builds on this observation and proposes to first pretrain the anticipation model on a large corpus of text-based recipes, \ie, Recipe1M+ \cite{Recipe1M} to acquire knowledge about the recipe domain, and then fine-tune the model on visual data.
This line of work effectively alleviates the video annotation problem, however, these works only predict a single future realization, and thus it does not take into account all the variability present in the recipes. 
For example, given the task of making a salad, and assuming the first three observed step are: \textit{Chop Vegetable}, \textit{Add Tomatoes}, \textit{Add Cucumber}, the plausible next step can be: \textit{Add Olive Oil} or \textit{Add Salt and Pepper} (for those who like more seasoning) or simply \textit{Serve}. 
This simple example highlights that the task's output is, in fact, multi-modal. This observation suggests that a good future step anticipation model must be able to predict diverse and plausible future realizations.
Moreover, it is known that in a multi-modal setup using a model that outputs a single prediction (as done by previous work~\cite{sener.etal.2022,yang.etal.2022}) may harm the performance~\cite{zhang2016colorful} even further by producing unrealistic samples that ``fall between'' the true modes.

In this work, we embrace the uncertainty inherent in the task of future anticipation and propose to learn a \underline{Ge}nerative \underline{P}rocedure \underline{S}tep \underline{An}ticipation model (\textsc{GePSAn}), that captures the distribution of possible next steps and allows to sample multiple plausible outputs given an initial observation from video input. A summary of our proposed work is depicted in Figure~\ref{fig:teaser}.
To achieve this goal, we design a model that consists of three modules: 1) a modality encoder, that ingests a sequence of previous instruction step observations (in either text or video format), 2) a generative recipe encoder, that, given the observation history, proposes the next plausible instruction vector, and 3) an instruction decoder, that transforms the next step prediction (given by the recipe encoder) into rich natural language.
The core component of the model is the generative recipe encoder; it combines the benefits of the transformer model~\cite{vaswani2017attention} (to process long input sequences) and Conditional Variational AutoEncoder (CVAE) (to capture the uncertainty inherent to the task) and can produce multiple plausible alternatives for the next step in a procedure.
Another key element of the pipeline is the input encoder; in contrast to the previous works that learn it from scratch, we adapt a pretrained video-language feature extractor~\cite{univl} to serve as our encoder.
Since the encoder has been trained to map video and text into a common embedding space, our model, trained only on the recipe text corpus, can generalize to future step anticipation from video zero-shot, without any finetuning.

\paragraph{Contributions.} Our contributions are twofold:
\begin{itemize}
    \item We propose \textsc{GePSAn}, a new generative model for future step anticipation that captures the uncertainty 
    inherent in the task of next step prediction.
    \item We show that \textsc{GePSAn}, only trained using text recipes, can generalize to video step anticipation zero-shot, without finetuning or adaptation.
\end{itemize}
Thanks to that, we achieve state-of-the-art results in next step anticipation from video on YouCookII~\cite{YC2} and show that our model can generate diverse plausible next steps, outperforming the baselines in modelling the distribution of next steps.

%% file: Sections/related_work.tex

\paragraph{Procedure planning and future anticipation.}

Previous work on future anticipation \cite{farha.etal.2020,furnari.2019,ng.fernando.2020} and procedure planning in video \cite{bi.etal.2021,chang.etal.2020,sun.etal.2021} is mainly based on the visual modality and relies on strong visual supervision.
Moreover, action anticipation is often considered as a classification problem, where the task is to predict a future action label from a predefined closed action set, \eg, \cite{girdhar2021anticipative,xu2022learning,gong2022future,nawhal2022rethinking,zhong2023anticipative,zatsarynna2021multi}.
Unlike most previous work, we rely on weak supervision from the language modality, and predict future actions in rich natural language. 
This allows us to transfer knowledge from a large scale text data to the visual domain, while only requiring a small text-aligned video data.

A few recent studies draw on language
 instructions as a source of weak supervision to perform 
 future anticipation for procedural activities \cite{sener.yao.2019,sener.etal.2022,yang.etal.2022,ghosh.etal.2023}.
Given a portion of an instructional video, these models predict plausible future actions, expressed using natural language.
Whereas the model of \cite{sener.yao.2019,sener.etal.2022} works by re-using the text recipe model parameters and fitting a visual encoder to the rest of the pretrained model, the models of \cite{yang.etal.2022,ghosh.etal.2023} transfer textual knowledge to the task of visual action anticipation via knowledge distillation.
Our work is most similar to that of~\cite{sener.yao.2019,sener.etal.2022}, but we offer a number of improvements. First, we design a modern transformer-based architecture and present an improved training objective with complementary loss functions.
Second, for efficient transfer learning across modalities, we leverage single-modality (language and video) encoders that are jointly trained for cross-modality alignment. Importantly, thanks to this design choice, we not only show improved predictions 
compared to relevant previous work of~\cite{sener.etal.2022,sener.yao.2019}, but also present a new benchmark in zero-shot cross-modality transfer with competitive performance.
Further, we consider diversity in future prediction task to capture the inherent uncertainty in future anticipation, which is overlooked in most related prior work on future anticipation from video \cite{sener.etal.2022,sener.yao.2019}.  
Notably, recent work on probabilistic procedure planning \cite{p3iv.2022} also explicitly tackles uncertainty. However, unlike this work, which is conditioned on the start and end of the procedure, we model uncertainty in a more challenging setting, where we only observe the start of the procedure, thereby making an approach modeling uncertainty even more relevant. 

\paragraph{Visual-textual representation learning.}
The ability to learn with minimal supervision is becoming increasingly important, and as such
recent work focuses on the complementarity across the visual and textual modalities as an inexpensive source of supervision \cite{desai.2021,radford.etal.2021,p3iv.2022,sun2019videobert}.
To further enable the use of cross-modal supervision, a large body of work uses \textit{aligned} multimodal data for learning rich representations that can be adapted for downstream tasks with minimal finetuning on task-specific annotations \cite{univl,miech.etal.2020,xu2021videoclip}. Such methods learn multiple single-modality encoders, each producing features aligned with the other modalities. We leverage the multimodal alignment offered by one such model - UniVL~\cite{univl} - to facilitate cross-modal transfer learning. For our initial text-only model, we adopt a frozen pretrained UniVL text encoder. Later, for transfer learning to videos, we replace the above encoder with a frozen pretrained UniVL video encoder. Importantly, our model design is not tied to this specific encoder but can readily be adapted to leverage stronger modality encoders.


\paragraph{Modeling uncertainty and diversity.}
Many previous works explicitly model uncertainty inherent to the task and leverage it to produce diverse output. Recent works \cite{petrovich21actor,yuan2020dlow} propose VAE based methods for diverse human motion synthesis. Other work proposes an RNN augmented with a VAE for stochastic sequence modelling \cite{goyal.etal.2017} . They evaluate their approach on speech and sequential image generation. In the context of action anticipation, some approaches use VAE to predict diverse future actions for the objects in the static image in terms of pixel-wise motion, \eg, \cite{walker2016uncertain}, while others use conditional GAN for long-term discrete action label anticipation \cite{zhao.wildes.2020}.
Unlike the existing works, we tackle diversity for future action prediction in natural language.
Thus, we find inspiration from various approaches for diverse dialogue modelling.
DialogWAE~\cite{gudialogwae} is a Wasserstein autoencoder (WAE) based solution for dialogue modeling. 
SPACEFUSION~\cite{gao2019jointly} relies on the fusion between seq2seq model and VAE for diverse Neural Response Generation.
Knowledge-Guided CVAE~\cite{zhao.etal.2017} provides discourse-level diversity for open-domain conversations.
Similar to the above works, we adopt conditional VAE in our model to predict multiple plausible next steps in rich natural language.

%% file: Sections/approach.tex
\begin{figure*}[t]
    \centering
    \includegraphics[width=\textwidth]{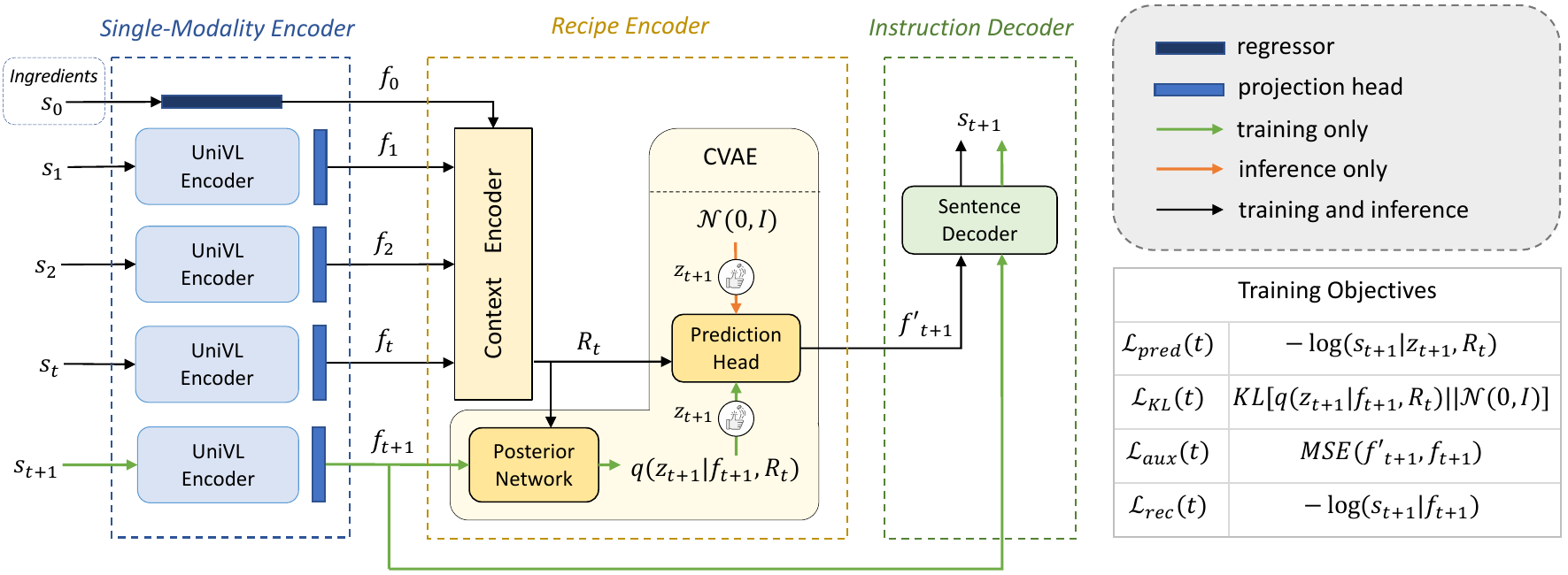}
    \caption{\textbf{Detailed view of our \textsc{GePSAn}.} Our \textsc{GePSAn} consists of three modules: a single-modality encoder (\ie, text or video encoder), a recipe encoder and an instruction decoder. \textbf{At training time:} A single-modality  encoder processes steps $s_{1:t+1}$ (text or video format) independently and produces features $f_{1:t+1}$. Next, given $f_{1:t+1}$, a recipe encoder reconstructs $f_{t+1}$ as follows. First, a context encoder predicts a context vector $R_t$ from $f_{1:t}$ and passes it to a CVAE. Then, in the CVAE, i) a posterior network predicts a posterior distribution $q(z_{t+1}|f_{t+1}, R_t)$ from $f_{1:t}$ and $R_t$, and ii) a prediction head reconstructs $f_{t+1}$ from $R_t$ and a sample $z_{t+1}\sim q(z_{t+1}|f_{t+1},R_t)$. We denote the reconstructed $f_{t+1}$ as $f'_{t+1}$. We pass the above $f'_{t+1}$ to the instruction decoder, which predicts $s_{t+1}$ in natural language. Additionally, the instruction decoder also decodes $f_{t+1}$ to predict $s_{t+1}$ in natural language. We train our model using the training objectives shown in the table on the right. \textbf{At inference time:} Given $s_{1:t}$ (text or video), a single-modal encoder predicts features $f_{1:t}$. Given $f_{1:t}$, a context encoder predicts a context $R_t$ and passes it to the CVAE. In the CVAE, i) we draw \textit{multiple} samples for $z_{t+1}$ from a Gaussian prior, and ii) we pass each $z_{t+1}$ and $R_t$ to a prediction head, which predicts multiple independent $f'_{t+1}$. Given multiple $f'_{t+1}$, a sentence decoder predicts diverse and independent alternatives for $s_{t+1}$.}
    \label{fig:arcitechture}
\end{figure*}

In this section, we formalize the problem of multi-modal future step anticipation, where the task is to predict \emph{multiple} plausible next steps (Sec.~\ref{sec:definition}). We then describe our solution - a generative future step prediction model (Sec.~\ref{sec:model}).
Sec.~\ref{sec:training} describes the proposed training objectives that we use to train a model  from a pure text-based corpus. Next, Sec.~\ref{sec:transfer} describes how to transfer the learned model to the video domain, with little fine-tuning or completely zero-shot. Finally, we provide implementation details in Sec.~\ref{sec:imp}.

\subsection{Problem definition}\label{sec:definition}
In this work we tackle the task of next step anticipation in procedural activities, \eg cooking, and propose to explicitly capture the multi-modal nature inherent to the task of future prediction. Specifically, given the first $t$ steps $s_{1:t}$ of a recipe, in video (or textual) format, our model outputs one or multiple ($k \geq 1)$ plausible options for the next procedure step, each expressed as a natural language sentence, \ie, $\{ s^{(1)}_{t+1}, \cdots, s^{(k)}_{t+1} \}$.
To better specify the prediction problem, we follow the prior work \cite{sener.etal.2022,sener.yao.2019,yang.etal.2022} and use the ingredient list as the $0$-th step, $s_0$.

\subsection{Model}\label{sec:model}
We illustrate our model in Figure \ref{fig:arcitechture}.
Our model consists three modules; namely, a single modality encoder, a recipe encoder and an instruction decoder.

\vspace{2mm}
\noindent\textbf{Single-modality (text or video) encoder.}
Given the observed instruction steps, $s_{0:t}$, in text or video domain, our single-modality encoders produce embeddings $f_{0:t}$.
Unlike prior work~\cite{sener.etal.2022,sener.yao.2019}, we use UniVL~\cite{univl} language and video encoders that were pretrained to embed the sentences or video clips into a common feature space.
Additionally, we augment UniVL (text or video) features with a learnable projection head, $P$, such that $f_t = P(\mathrm{UniVL}(s_t))$.

\vspace{2mm}
\noindent\textbf{Recipe encoder.}
The recipe encoder is the core of our model, it takes in a sequence of embeddings, $f_{0:t}$, corresponding to $t$ observed instruction steps, $s_{0:t}$, and outputs multiple plausible future step embeddings, $\{ f'^{(1)}_{t+1}, \cdots, f'^{(k)}_{t+1} \}$.
It consists of two components: a context encoder and a conditional Variational Auto Encoder (CVAE).

\vspace{1mm}
\noindent\textit{Context encoder.} The context encoder is implemented as a transformer~\cite{vaswani2017attention} block.
It aggregates past sentence embeddings $f_{0:t}$ into a single context vector $R_t$. 
To take only the past history into account, we use causal attention in our context encoder, that is, we only use $f_{0:t}$ to produce $R_{t}$.
During training, our context encoder observes all embeddings up to $t-1$ and produces $R_{0:t-1}$ simultaneously.

\vspace{1mm}
\noindent\textit{Conditional VAE.} Our CVAE consists of a posterior network and a prediction head. During training, the posterior network ingests the concatentation of the context vector, $R_t$, (\ie, conditional input) and the next sentence embedding, $f_{t+1}$, and predicts a posterior $q(z_{t+1}|f_{t+1}, R_t)$. We then sample a latent, $z_{t+1} \sim q(z_{t+1}|f_{t+1}, R_t)$, concatenate it with the context vector, $R_t$, and pass them to the predication head to predict the embedding of the next instruction $f'_{t+1}$.
During training, we minimize KL divergence between the predicted posterior $q(z_{t+1}|f_{t+1}, R_t)$ and a standard Gaussian prior with zero mean and unit variance. Thus, at inference, we discard the posterior network, sample $z_{t+1}$ from $\mathcal{N}(0, I)$, and follow the same steps as described above.
This way, by using a CVAE on top of the context encoder, we essentially learn a distribution of the next steps, conditioned on the observed step history.
Another advantage of the CVAE framework is fast sampling at test time.

\vspace{2mm}
\noindent\textbf{Instruction decoder. }
Given a predicted embedding $f'_{t+1}$, our instruction decoder decodes the next step in natural language. We implement the instruction decoder as a simple LSTM (not a transformer) as it demonstrated better validation results.

At inference, we sample one or multiple ($k \geq 1$) $z^{(k)}_{t+1}$ from $\mathcal{N}(0, I)$ and combine it with a given context $R_t$.
Subsequently, our CVAE predicts multiple embeddings $\{ f'^{(1)}_{t+1}, \cdots, f'^{(k)}_{t+1} \}$ and our sentence decoder decodes each embedding into a separate and independent alternative for the next step.

\subsection{Training objectives}~\label{sec:training}
Our training objective combines three losses that help \textsc{GePSAn} capture the probability distribution of next steps, provide good sentence decoding, and stabilize training.
\paragraph{Conditional evidence lower bound.}
Conditional Evidence Lower Bound (or conditional ELBO) is the loss used to train the CVAE and is mostly responsible for capturing the multi-modal distribution associated with the task of next step prediction.
The conditional ELBO used to train our model can be expressed as follows:
\begin{align}
    \mathcal{L}_{ELBO}(t) &= \mathcal{L}_{pred}(t) + \beta \mathcal{L}_{KL}(t).\\
    \mathcal{L}_{pred}(t) &= - \sum_{j=1}^{M_{t+1}} \log p(w^j_{t+1} | w^{j'<j}_{t+1}, z_{t+1}, R_t), \label{eq:Lpred}\\
    \mathcal{L}_{KL}(t) &= \mathbf{KL} [q(z_{t+1}|f_{t+1}, R_t)||\mathcal{N}(0,I)],
    \label{eq:elbo}
\end{align}
where $z_{t+1}\sim \mathbb{E}_{q(z_{t+1}|f_{t+1}, R_t)}$ and $w^j_{t+1}$ is the $j^{th}$ word out of total $M_{t+1}$ words in the $(t+1)^{th}$ sentence. To avoid posterior collapse in the early epochs, we introduce $\beta$ coefficient for the KL divergence in the above objective and anneal $\beta$ linearly~\cite{bowman2016generating}.

\paragraph{Auxiliary objective.}
Previous works \cite{zhao.etal.2017, goyal.etal.2017} suggest that an auxiliary loss is essential to train a CVAE along with a seq2seq model. Thus, we introduce an additional auxiliary loss, 
\begin{align}
    \mathcal{L}_{aux}(t) = \mathrm{MSE}(f'_{t+1}, f_{t+1}).
    \label{eq:regularizer}
\end{align}
Note that we compute gradient of the above objective with respect to $f'_{t+1}$. The embeddings $f_{t+1}$ act as a target only. Notably, the auxiliary loss also simplifies the sentence decoding process (\ie, Eq. \ref{eq:Lpred}) by: i) compelling our CVAE to reconstruct the embeddings $f'_{t+1}$ given $[z_{t+1};R_t]$, and ii) allowing our sentence decoder to decode the $(t+1)^{th}$ sentence from $f'_{t+1}$.

\paragraph{Sentence reconstruction objective.}
While 
the auxiliary loss is designed to simplify the decoding process, initially, it is still difficult for the sentence decoder to predict the next 
sentence since $f'_{t+1}$ is not yet learnt. Therefore, we also train our sentence decoder to reconstruct the individual sentences from their projected UniVL embeddings, \ie, $f_{t+1}$. Note that $f_{t+1}$ is a more stable input for the decoder compared to $f'_{t+1}$. The reconstruction objective is,
\begin{align}
    \mathcal{L}_{rec}(t) = - \sum_{j=1}^{M_{t+1}} \log p(w^j_{t+1} | w^{j'<j}_{t+1}, f_{t+1}).
    \label{eq:decode}
\end{align}
Our final training objective is:
\begin{align}
    \mathcal{L} &= \sum_{t=0}^{T-1} \big\{\mathcal{L}_{ELBO}(t) + \alpha \mathcal{L}_{aux}(t) + \gamma \mathcal{L}_{rec}(t) \big\}.
    \label{eq:final_loss}
\end{align}
where $\alpha$ and $\gamma$ are hyperparameters used to balance the training objectives.

\paragraph{Pretraining domain.}
Due to its flexible design, \textsc{GePSAn} can ingest instruction step representations $s_{0:t}$ in the form of text or video.
While our final objective is to do next step prediction purely from video, the size of annotated video-based cooking datasets does not allow us to train such a model from scratch.
Thus, we follow prior work~\cite{sener.etal.2022,sener.yao.2019} and pretrain \textsc{GePSAn} on a large corpus of text-only recipes.
That is, given a sequence of step sentences as input, our pretraining objective is to model the next step distribution from textual input only, using the training pipeline and the final loss (in Eq.~\ref{eq:final_loss}) described above.
After this pretraining stage, the model can be adapted to take video as input, using a modest amount of fine-tuning or completely zero-shot.

\subsection{Transfer learning}\label{sec:transfer}
After the model has been pretrained on the text corpus, as described above, we adapt the model to accept video snippets as input.
To do so, we replace the frozen UniVL sentence encoder with the frozen UniVL video encoder.
Since the UniVL video and text features are aligned by design (\ie, they live in the same embedding space), after the above switch, our model readily offers strong future step anticipation performance, without further fine-tuning or adaptation.
We refer to this setting as \textit{zero-shot modality transfer}.
Optionally, to further boost the step anticipation performance, we can finetune \textsc{GePSAn} on a small annotated video dataset, as done in previous work~\cite{sener.etal.2022,sener.yao.2019} by default.
It is important to note that the finetuning stage is essential for the previous methods to work, and is only optional in our case, as we can already perform future step anticipation from video without any finetuning with competitive results as we later demonstarte in the experiments section.

\subsection{Implementation details}\label{sec:imp}
For the loss hyperparameters, we set $\alpha=3$ for Recipe1M+, and  $\alpha=1$ for YouCookII. We set $\gamma=1$ in all cases. For Recipe1M+, we set $\beta=0.2$ with linear KL annealing to reach that value in $100,000$ steps. $\beta$ is set to $0.1$ for YouCookII. We use the Adam optimizer~\cite{adam.2015} with a learning rate of $0.0001$, weight decay of $0.01$, and a one epoch linear warm-up. The batch size is set to $50$.
Regarding the architecture, we used a 3-layer residual block for the UniVL projection head, and a one-layer MLP for the ingredients regressor that takes as input a one-hot vector of ingredients (the number of ingredients is $3,769$). The context encoder is a 6-layer transformer with an input dimension of $512$ and $8$ heads. The posterior network and the prediction head are both 3-layer MLPs, with the latent variable $z$ having a dimension of $1024$. The instruction decoder is a 3-layer LSTM with a hidden size of $512$ and a word embedding size of $256$.

%% file: Sections/experiments.tex
In this section, we evaluate the performance of \textsc{GePSAn}, our proposed approach, on the task of future step anticipation, make a comparison to relevant baselines, and perform an ablation study of the proposed model components.
We begin by detailing the experimental setup and adopted evaluation metrics given our newly proposed view, which takes the multi-modal nature of the task into account (Sec.~\ref{exp-setup}). We then evaluate our model on the main task of future step anticipation from video input (Sec.~\ref{results}). We finally show the role of pretraining on a large text-based dataset, highlight the flexibility of our model that can take video or text as input, and demonstrate the role of each component in our training objective (Sec.~\ref{ablation}).

\subsection{Experimental setup}\label{exp-setup}
\begin{table*}[t!]
\centering
\begin{tabular}{l|l||ll|lll||ll|lll}
\toprule
& \textbf{Model} & \multicolumn{5}{c||}{\textbf{\textit{Unseen Split}}} & \multicolumn{5}{c}{\textbf{\textit{Seen Split}}} \\
 & &\textsc{ING} & \textsc{VERB} &\textsc{B1} & \textsc{B4} & \textsc{MET} &\textsc{ING} & \textsc{VERB} &\textsc{B1} & \textsc{B4} & \textsc{MET} \\
\toprule
 \multirow{2}{*}{\parbox{3cm}{\textbf{Text~$\rightarrow$~Video Zero-shot Transfer}}} & \textsc{GePSAn} \single\  & 16.5 & 24.1 & 23.0 & 2.2 & 8.3 & - & - & - & - & -\\
 & \textsc{GePSAn} \multiple\ \setrow{\bfseries}& 30.0 & 28.7 & 31.4 & 3.7 & 11.6 & - & - & - & - & - \\
 \midrule
 \midrule
\multirow{4}{*}{\parbox{2cm}{\textbf{Finetuning on Video}}} & \textsc{Baseline} \single\ & 16.8 & 26.9 & 25.1 & 3.1 & 9.2 & 19.6 & 27.5 & 25.8 & 4.0 & 9.8 \\
 & \textsc{GePSAn} \single\ \setrow{\bfseries}& 21.5 & 29.9 & 27.6 & 4.8 & 10.8 &  25.6 & 30.8 & 28.9 & 5.8 & 11.8 \\
\cline{2-12}
 & \textsc{Baseline} \multiple\ $^\diamond$ & 27.8 & 31.6 & 33.1 & 4.4 & 12.3 & 32.2 &34.2 &35.0 &5.9 & 13.7\\
 & \textsc{GePSAn} \multiple\ \setrow{\bfseries}& 31.6 &  37.8 & 35.6 & 7.9 & 14.5 &  36.7 & 38.4 & 37.1 & 9.3 & 15.7 \\
\bottomrule
\end{tabular}
\caption{\textbf{YouCookII future anticipation from video input.}
We report results for two settings: (top) zero-shot text-to-video modality transfer and (bottom) finetuning on video modality. For each setting, we compare our results with the baseline~\cite{sener.etal.2022} results (when available) for single~\single\ and multiple~\multiple\ next step prediction. To achieve single and multiple predictions, we evaluate \textsc{GePSAn} using latent $z_{t+1}=0$ (\ie, mean of a Gaussian prior) and five random $z_{t+1}\sim \mathcal{N}(0,I)$, respectively.  $^\diamond$We use \textit{Nucleus sampling}~\cite{holtzmancurious} to achieve multiple predictions from the deterministic baseline. Further, we present comparison for recipe-types unseen and seen in the training split.
}
\label{res:1}
\end{table*}
\begin{table*}[t]
\centering
\begin{tabularx}{\textwidth}{l|l||ll|lll||ll|lll}
\toprule
& \textbf{Model} & \multicolumn{5}{c||}{\textbf{\textit{Unseen Split}}} & \multicolumn{5}{c}{\textbf{\textit{Seen Split}}} \\
 & &\textsc{ING} & \textsc{VERB} &\textsc{B1} & \textsc{B4} & \textsc{MET} &\textsc{ING} & \textsc{VERB} &\textsc{B1} & \textsc{B4} & \textsc{MET} \\
\toprule
\multirow{4}{*}{\parbox{2cm}{\textbf{Zero-shot Dataset Transfer}}} & \textsc{Baseline} \single\ & 22.9 & 29.1 & 25.8 & 3.0 & 10.0 & - & - & - & - & - \\
  & \textsc{GePSAn} \single\ & 20.0 & 28.3 & 24.7 & 3.1 & 9.4 & - & - & - & - & -  \\
  \cline{2-12}
   & \textsc{Baseline} \multiple\ $^\diamond$ & 27.1 & 31.3 & 29.3 & 2.5 & 11.4 & - & - & - & - & - \\
 & \textsc{GePSAn} \multiple\ \setrow{\bfseries}& \textbf{33.3} & \textbf{33.7} & \textbf{32.4} & \textbf{4.7} & \textbf{15.2} & - & - & - & - & - \\
 \midrule
 \midrule
\multirow{4}{*}{\parbox{2cm}{\textbf{Finetuning on the Target Dataset}}} & \textsc{Baseline} \single\ & 26.9 & 31.8 & 30.6 & 6.6 & 12.2 & 29.1 & 32.9 & 31.0 & 7.3 & 12.8 \\
 & \textsc{GePSAn} \single\ \setrow{\bfseries}& 28.9 & 33.7 & 33.0 & 7.2 & 13.2 & 32.7 & 35.2 & 35.0 & 8.5 & 14.4 \\
\cline{2-12}
 & \textsc{Baseline} \multiple\ $^\diamond$ & 38.4 & 38.8 & 39.3 & 8.6 & 15.8 & 40.0 & 39.2 & 39.6 & 8.8 & 16.1 \\
 & \textsc{GePSAn} \multiple\ \setrow{\bfseries}&\textbf{41.7} & \textbf{42.9} & \textbf{41.4} & \textbf{11.0} & \textbf{17.3} & \textbf{44.6} & \textbf{43.7} & \textbf{43.0} & \textbf{12.3} & \textbf{18.4} \\
\bottomrule
\end{tabularx}
\caption{\textbf{YouCookII future anticipation from text input.}
We report results for two settings: (top) zero-shot dataset transfer and (bottom) finetuning on the target data. For each setting, we compare our results with the baseline~\cite{sener.etal.2022} results (when available) for single~\single\ and multiple~\multiple\ next step prediction. To achieve single and multiple predictions, we evaluate \textsc{GePSAn} using latent $z_{t+1}=0$ (\ie, mean of a Gaussian prior) and five random $z_{t+1}\sim \mathcal{N}(0,I)$, respectively. $^\diamond$We use \textit{Nucleus sampling}~\cite{holtzmancurious} to achieve multiple predictions from the deterministic baseline. Further, we present comparison for recipe-types unseen and seen in the training split.
}
\label{res:youcookii_textual}
\end{table*}
\paragraph{Datasets.}
For the text-only pretraining stage, we follow previous work and take advantage of a large text-based recipe dataset. Specifically, we use the publicly available Recipe1M+ dataset \cite{Recipe1M}, which contains over one million cooking recipes to pretrain our model. Then, given that the main target of the proposed approach is future step anticipation from \emph{video} input, we follow previous work \cite{sener.yao.2019} and evaluate on the YouCookII dataset \cite{YC2}, a video-based cooking dataset, (\ie, we use Recipe1M+ for learning procedural knowledge from text, and showcase transfer learning to visual domain on YouCookII). YouCookII consists of 2000 long untrimmed videos (in 3rd person viewpoint) from 89 cooking recipes. Each video is associated with an ordered list of steps describing the recipe being performed in free form natural language, together with start and end times of each step in the video. Notably, while previous work also evaluated on the Tasty video dataset \cite{sener.yao.2019}, it is not used in this work due to copyright limitations. 

\begin{figure*}[t]
    \centering
    \includegraphics[width=0.85\textwidth]{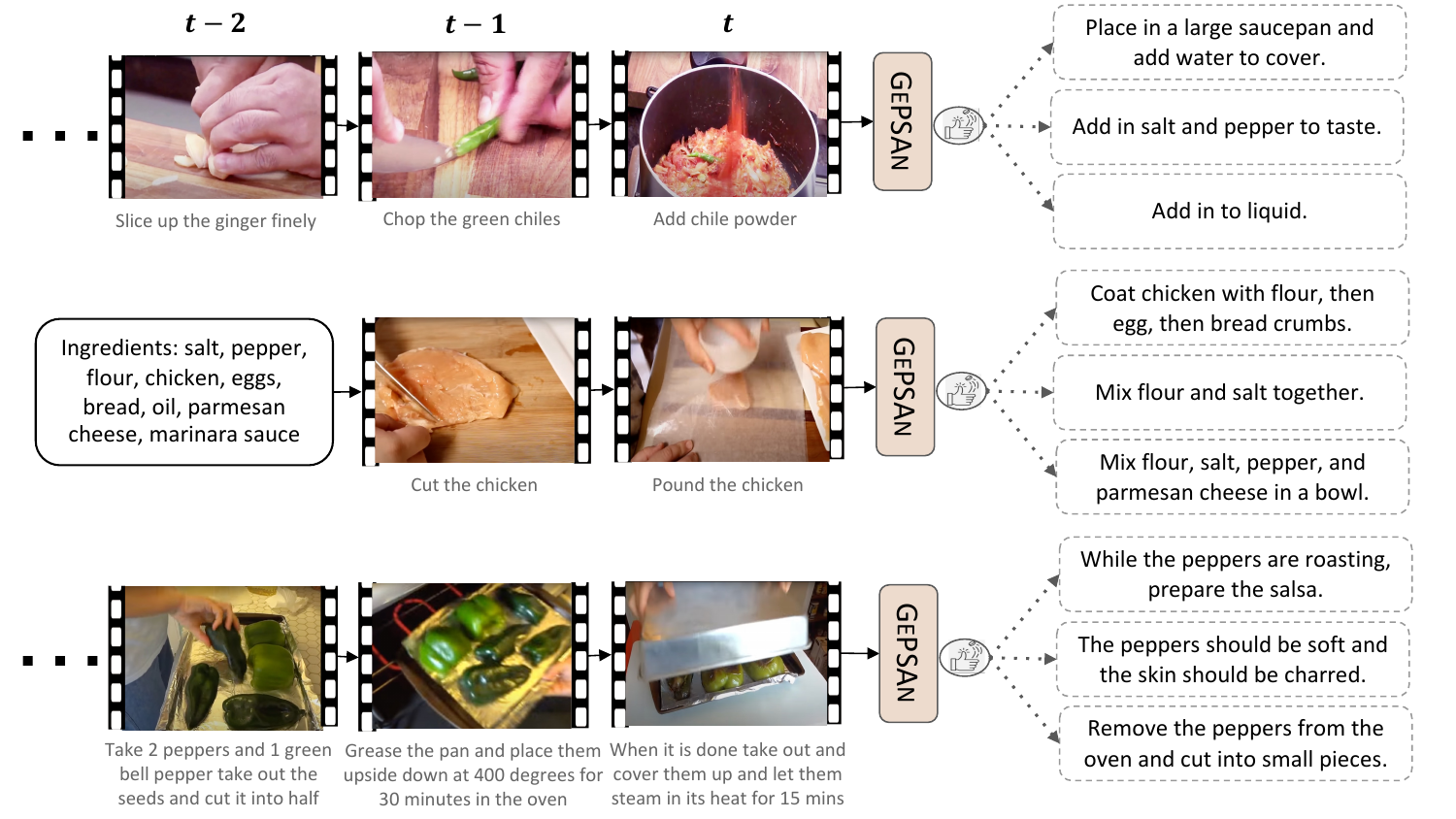}
    \caption{\textbf{Qualitative results on the Text~$\rightarrow$~Video Zero-shot Transfer without any finetuning on videos.} Please note that only the video is passed to the model as input, the text here is provided for context.}
    \label{fig:qual}
\end{figure*}
\paragraph{Evaluation metrics.}
Our model predicts next steps in free-form natural language, therefore, we follow standard protocol \cite{sener.yao.2019, sener.etal.2022} and evaluate using sentence matching scores, including: BLEU1 (B1), BLEU4 (B4) and METEOR (MET). Notably, we use the standard corpus-level calculation method for the BLEU~\cite{bleu} and METEOR scores, while previous work \cite{sener.yao.2019, sener.etal.2022} used the average of sentence-level BLEU and METEOR scores; see supplement for a detailed discussion and the complete set of results using both methodologies.

We also calculate the recall on the set of ingredients (ING) and verbs (VERB) included in the ground truth sentence, \ie, we calculate the ratio of the verbs and ingredients in the ground truth predicted by the model. Note that the recalls on ING and VERB are stronger indicators of model performance as they highlight the diversity of the predicted actions rather than the diversity in the sentence styles.

Importantly, unlike previous work, our approach can predict either \emph{multiple} \multiple plausible next steps or a \emph{single} \single next step (\ie, by setting the latent $z_{t+1}$ to the maximum likelihood sample from the latent prior distribution, $\mathcal{N}(0, I)$, which happens to be a zero vector). To evaluate our approach in the \emph{multiple} setting using the same evaluation metrics described above, we have to select just one out of $k$ predicted next steps, since we only have one ground-truth in the dataset.
To do so, we pick the predicted step that is closest to the ground truth sentence using the Jaccard Similarity (Intersection over Union).
Precisely, we treat the words in each sentence as a Bag of Words (BoW) and calculate the Jaccard similarity between each predicted sentence and the ground truth.
Notably, we elect to use Jaccard similarity here as it is a well accepted metric for comparing two sentences, but any alternative metric for comparing sentences can be used for this step.
Intuitively, this inference procedure is meaningful because the model samples multiple plausible next steps, only one of which is contained in the labels, so the matching step described above associates the ground-truth with the closest predicted sample. $k$ is set to $5$ in all the experiemnts. See supplement for the impact of increasing $k$ on the results.

\paragraph{Baselines.} We use previous state of the art for future step anticipation from cooking videos \cite{sener.etal.2022}\footnote{We use the code shared with us by the authors.} as our main \textsc{baseline} and compare our \emph{single} \single prediction setting directly to it. For our \emph{multiple} \multiple next step prediction setting, there is no directly comparable baseline, to the best of our knowledge. Therefore, we augment our main \textsc{baseline} with a simple approach to generate multiple next steps. Specifically, we replace the deterministic greedy approach used in the decoder of the \textsc{baseline}, with the Nucleus sampling method \cite{holtzmancurious} to generate $k$ alternatives for the next step.

\subsection{Results}\label{results}
\paragraph{YouCookII video-based future anticipation.}
As previously mentioned, the main target of this work is future step anticipation given video input.
Note that we evaluate both the \emph{single} (\ie deterministic) and \emph{multiple} (\ie generative) prediction versions of our approach in two main settings; namely, \textbf{(i)} text $\rightarrow$ video zero-shot transfer and \textbf{(ii)} with video finetuning.
In the zero-shot modality transfer setting, we use our model pretrained on the text-based Recipe1M+ dataset, and directly replace the textual input with visual input from the YouCookII dataset, and use the UniVL video encoder instead of the text encoder. 
In the modality finetuning setting, we further finetune the model, except for the pretrained UniVL encoder, on the training split of YouCookII.

In addition to the variations in terms of training settings, we follow previous work \cite{sener.etal.2022} and also assess our model on two different splits of the YouCookII dataset; namely, \textbf{(i)} unseen split, containing recipes never seen during training, and \textbf{(ii)} seen split, containing only seen recipes \cite{sener.etal.2022}.

Table~\ref{res:1} summarizes our main results on the YouCookII dataset. The results of both variants of our model (\ie, \single vs. \multiple) compared to the baseline speak decisively in favor of our approach, where we outperform the baseline under all settings in all the considered metrics. Notably, comparing our model to the baseline in the zero-shot modality transfer setting is \emph{not} possible, as the baseline relies on training a new model for the visual modality. In contrast, we judiciously use modality encoders that were pretrained, in unsupervised settings, such that visual and language representations share a common embedding space. This strategy allowed us to focus on pretraining a stronger recipe encoder, which is reflected in results reported in Table~\ref{res:1}. 

Importantly, comparing the \emph{multiple} \multiple prediction to \emph{single} \single prediction settings, highlights the importance of modeling the uncertainty inherent to the task of future anticipation, where a model yielding multiple plausible outputs better captures possible future steps, as evidenced by \multiple results always outperforming \single results. 
Figure~\ref{fig:qual} illustrates some qualitative results from \textsc{GePSAn} \multiple, thereby further validating the relevance of the multiple plausible future step predictions. Obtaining corresponding quantitative results to better quantify the plausibility of the generated next steps is unfortunately not possible in the absence of a dataset capturing multiple feasible ground truths, which would allow calculating precision. Curating such a dataset is outside the scope of this paper.

\paragraph{YouCookII text-based future anticipation.} While our main goal is video-based future step anticipation, for the sake of completeness we also evaluate our model on text-based future anticipation. The results of predicting \emph{multiple} plausible next steps in Table~\ref{res:youcookii_textual} further confirm the superiority of our approach and the importance of capturing the uncertainty inherent to the task of future prediction. As expected, the results with text-based input are better, compared to the visual-based input, as there is no modality change in these settings. However, although there is no change of modality here, the results before and after finetuning indicate that there is a difference in the distributions of the two datasets. More generally, these results highlight the flexibility of our model that can readily use either textual or visual input in zero-shot settings, unlike previous work \cite{sener.etal.2022}.

\subsection{Ablations}\label{ablation}
 
\paragraph{Contribution of the different training objectives.} We evaluate the role
of each loss component by gradually removing each objective. The results in Table~\ref{res:contribution} confirm the pivotal role of the auxiliary loss, $\mathcal{L}_{aux}$, to train the CVAE as mentioned in Sec.~\ref{sec:training}. The prediction loss, $\mathcal{L}_{pred}$, also plays an important role in boosting performance. Note that removing the KL divergence, $\mathcal{L}_{KL}$, leads to model divergence. Finally, while the results in Table~\ref{res:contribution} suggest that the reconstruction loss, $\mathcal{L}_{rec}$, does not contribute much to the model, during our experiment we noted that it plays an important role earlier during training and helps in faster and smoother convergence.
\begin{table}[t]
\centering
\begin{tabular}{@{}C{3cm}||C{0.5cm}C{0.9cm}|C{0.5cm}C{0.5cm}C{0.8cm}@{}}
\toprule
\textbf{Model} & \multicolumn{5}{c}{\textbf{Recipe1M+} (Textual)}\\
 &\textsc{ING} & \textsc{VERB} &\textsc{B1} & \textsc{B4} & \textsc{MET} \\
\toprule
\textsc{GePSAn} \single\ & 27.2 & 28.5 & 25.9 & 7.5 & 11.2 \\
\textsc{GePSAn} \multiple\ & 37.2 & 36.2 & 32.2 &  10.7 & 14.6 \\
\midrule
w/o $\mathcal{L}_{aux}$ \single\ & 26.8 & 27.9 & 23.9 & 7.4 & 10.9 \\
w/o $\mathcal{L}_{aux}$ \multiple\ & 29.4 & 29.2 & 25.4 & 8.2 & 11.6 \\
\midrule
w/o $\mathcal{L}_{pred}$ \single\ & 25.7 & 29.0 & 25.9 & 5.2 & 11.0 \\
w/o $\mathcal{L}_{pred}$ \multiple\ & 34.0 & 35.4 & 33.2 & 7.7 & 13.8 \\
\midrule
w/o $ \mathcal{L}_{rec}$ \single\ & 27.7 & 28.3 & 25.7 & 7.3 & 11.0 \\
w/o $ \mathcal{L}_{rec}$ \multiple\ & 36.6 & 36.5 & 32.2 & 10.8 & 14.6 \\
\bottomrule
\end{tabular}
\caption{\textbf{Ablation Study for the text-based future anticipation on Recipe1M+.} We assess contribution of the individual training objectives during the model pretraining phase. We report results for single~\single\ and multiple~\multiple\ next step prediction. To achieve single and multiple predictions, we evaluate \textsc{GePSAn} using latent $z_{t+1}=0$ (\ie, mean of a Gaussian prior) and five random $z_{t+1}\sim \mathcal{N}(0,I)$, respectively.}
\label{res:contribution}
\end{table}


\paragraph{Recipe1M+ pretraining.}
Additionally, we include a comparison of the pretraining phase performance on the Recipe1M+ text dataset in Table~\ref{res:recipe1m}.
In the \emph{single} \single prediction setting, our performance is on-par, or slightly sub-par, with the \textsc{baseline}, which suggests that when training on a large textual dataset, it might be beneficial to learn the text encoder from scratch instead of using the pre-trained UniVL encoder (though the opposite is true for testing on video).
However, \textsc{GePSAn} outperforms the baseline in the \multiple prediction setting even with the sub-optimal text encoder, which shows that our model can capture the multi-modal nature of the task under such settings as well.
Notably, if the model is trained from scratch on YouCookII, the performance collapses, which shows the importance of the pretraining phase given data scarcity in the video domain, and the difficulty of training a generative model on such a small dataset.

\begin{table}[t]
\centering
\begin{tabular}
{l|| ll | lll}
\toprule
 \textbf{Model} & \textsc{ING} & \textsc{VERB} &\textsc{B1} & \textsc{B4} & \textsc{MET} \\ 
 \midrule
\textsc{Baseline} \single\ & 27.0 & 29.4 & 24.1 & 7.8 & 11.3 \\
\textsc{GePSAn} \single\ & 27.2 & 28.5 & 25.9 & 7.5 &11.2 \\
\midrule
\textsc{Baseline} \multiple\ $^\diamond$ & 34.7 & 34.6 & 31.7 & 9.4 & 14.2 \\
\textsc{GePSAn} \multiple\  & \textbf{37.2} & \textbf{36.2} & \textbf{32.2} &  \textbf{10.7} &\textbf{ 14.6} \\
\bottomrule
\end{tabular}
\caption{\textbf{Text-based future anticipation results on Recipe1M+.} We compare our results with the baseline~\cite{sener.etal.2022} results for single~\single\ and multiple~\multiple\ next step prediction. To achieve single and multiple predictions, we evaluate \textsc{GePSAn} using latent $z_{t+1}=0$ (\ie, mean of a Gaussian prior) and five random $z_{t+1}\sim \mathcal{N}(0,I)$, respectively. $^\diamond$We use \textit{Nucleus sampling}~\cite{holtzmancurious} to achieve multiple predictions from the deterministic baseline.}
\label{res:recipe1m}
\end{table}

%% file: Sections/conclusion.tex
In this work, we have addressed the problem of next step prediction from instructional videos (focusing on cooking activities).
In particular, we have proposed \textsc{GePSAn}, a generative next step prediction model that concentrates on capturing the uncertainty inherent to the task of future step anticipation, which was largely overlooked in previous work tackling this task in realistic open-world setting (\ie, not relying on a predefined closed set of step labels). In addition, we showed that \textsc{GePSAn} can effectively capture multiple feasible future realizations, and outperforms existing baselines on video anticipation, with or without domain-specific adaptation, \ie, zero-shot, thanks to the judicious use of aligned modality representation. We hope that this work will open up new avenues for future research that automatically considers multiple possible future realizations in open world next step prediction, with datasets and metrics that better support evaluation under these settings.


%% file: Sections/supplementary.tex
\subsection{Adding Cross Distillation Loss to Baseline}
Similar to \cite{sener.yao.2019, sener.etal.2022}, Yang \etal\cite{yang.etal.2022} learns to predict the next step in the recipe from a cooking video by transferring knowledge from the textual domain to visual domain. Motivated by knowledge distillation ~\cite{hinton2015distilling}, they use the textual future prediction model as teacher, while training a student model on videos. They call their method cross-modal contrastive distillation (CCD). They do this process after finetuning the textual model on the text of the video dataset (YouCookII in our case). For additional comparison, we extend our baseline from the main paper by training it using CCD on YouCookII. We report results in table \ref{tab:compTo31}. While the baseline method benefits from CCD, it still fails to match the performance of our \textsc{GePSAn}. The results indicate superiority of our generative approach over cross-modal distillation.
\begin{table*}[h!]
\centering
\resizebox{0.625\columnwidth}{!}{
\begin{tabular}{l| l||l|l|l|l|l}
\toprule
 & & ING & VERB & B1   & B4   & MET  \\
\midrule
\multirow{3}{*}{\parbox{4cm}{\textbf{Single~\single Prediction}}} & Baseline & 19.6 & 27.5 & 25.8 & 4.0  & 9.8  \\
 & CCD & 20.8 & 27.0 & 26.4 & 4.2  & 10.0 \\
& \textsc{GePSAn}                  & 25.6 & 30.8 & 28.9 & 5.8  & 11.8 \\
\midrule
\multirow{3}{*}{\parbox{4cm}{\textbf{Multiple \multiple Predictions}}} & Baseline$^\diamond$ & 32.2 & 34.2 & 35.0 & 5.9  & 13.7 \\
& CCD$^\diamond$ & 33.5 & 34.3 & 36.2 & 6.8  & 14.1 \\
& \textsc{GePSAn} & \bf{36.7} & \bf{38.4} & \bf{37.1} & \bf{9.3}  & \bf{15.7} \\
\bottomrule
\end{tabular}}
\caption{Future anticipation results on YouCookII Video after finetuning. CCD is our adaptation of \cite{yang.etal.2022} to the Baseline. $^\diamond$ We use \textit{Nucleus sampling}~\cite{holtzmancurious} to achieve multiple predictions from the deterministic baseline.}\label{tab:compTo31}
\end{table*}

\subsection{Impact of k}
\begin{figure}[h!]
    \centering
    \includegraphics[width=0.45\textwidth]{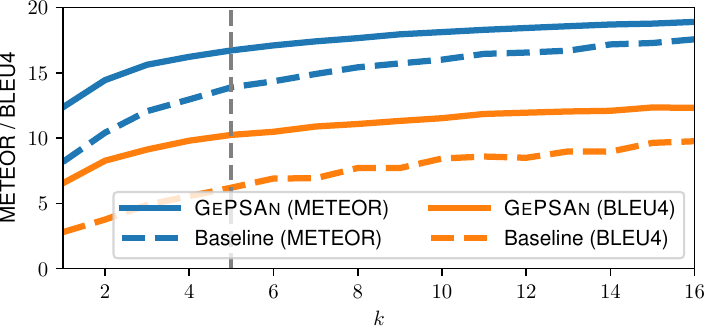}
    \caption{METEOR and BLEU4 with increasing 
    number of sampled predictions ($k$) on YouCookII videos after finetuning.}
    \label{fig:k_vs_meteor_bleu4}
\end{figure}

In Fig.~\ref{fig:k_vs_meteor_bleu4}, we show BLEU4 and METEOR scores of our \textsc{GePSAn} and the baseline with increasing value of $k$. Observe that the baseline fails to match the performance of \textsc{GePSAn}, suggesting that  simply increasing $k$ is not sufficient to improve performance if diversity is not meaningful. Higher performance of our model across $k$ shows that it indeed produces diverse yet meaningful predictions. Also the improvements with increasing $k$ plateaus early, indicating that \textsc{GePSAn} predicts steps highly relevant to the GT.

\label{sec:k_impact}

\subsection{Additional results with different evaluation methodologies}
In Tables~\ref{res:1} and~\ref{res:recipe1m} (main paper), we report our and the baseline results for BLEU1 (B1), BLEU4 (B4) and METEOR (MET) metrics computed using the standard corpus-level formula which uses the micro-averaged statistics before computing the corpus-level BLEU and METEOR scores~\cite{bleu}. However, the baseline results were originally reported by macro-averaging the sentence-level metrics~\cite{sener.etal.2022}. 
Here, in Table~\ref{res:recipe1m_reconc}, we report our results, as well as the reproduced baseline results, using both micro- and macro-averaging. We also present the original macro-averaged baseline results as reported in \cite{sener.etal.2022}.
We observe that the baseline macro-averaging results reproduced by us are similar to the ones reported in the original paper. Also, the micro-averaged  results follow the same trend as the  macro-averaged ones.

\begin{table*}[h!]
\centering
\resizebox{0.95\textwidth}{!}{%
\begin{tabular}
{l|| cc | cc | ccc | ccc | ccc}
\toprule
 \multirow{3}{*}{\textbf{Model}} & \multicolumn{2}{c}{ING} & \multicolumn{2}{c}{VERB} & \multicolumn{3}{c}{\textsc{B1}} & \multicolumn{3}{c}{\textsc{B4}} & \multicolumn{3}{c}{\textsc{MET}} \\ 
 \cmidrule(lr){2-10}
  & \textit{Micro} & \textit{Micro} & \textit{Micro} & \textit{Micro} & \textit{Micro} & \textit{Macro} & \textit{Macro} & \textit{Micro} & \textit{Macro} & \textit{Macro} & \textit{Micro} & \textit{Macro} & \textit{Macro} \\
&&\small{\cite{sener.etal.2022}}&&\small{\cite{sener.etal.2022}}&&&\small{\cite{sener.etal.2022}}&&&\small{\cite{sener.etal.2022}}&&&\small{\cite{sener.etal.2022}}\\
\midrule
\multicolumn{14}{c}{\textbf{YouCookII Video (Unseen Split)}} \\
  \midrule
\textsc{Baseline} \single\ &16.8&17.8&26.9&23.1& 25.1 & 22.4 & 20.6 & 3.1 &  0.6 & 0.84 & 9.2 & 9.9 & 9.5\\
\textsc{GePSAn} \single\ &21.5&-&29.9&-& 27.6 & 25.6 & - & 4.8 & 1.4 & - & 10.8 & 12.0 & -\\
\midrule
\textsc{Baseline} \multiple\ &27.8&-&31.6&-& 33.1 & 28.8 & - & 4.4 & 0.8 & - & 12.2 & 13.2 & -\\
\textsc{GePSAn} \multiple\  &\textbf{31.6}&&\textbf{37.8}&-& \textbf{35.6} & \textbf{33.0} & - &  \textbf{7.9} & \textbf{2.6}  & - & \textbf{14.5} & \textbf{16.0} & -\\
\midrule
\multicolumn{14}{c}{\textbf{YouCookII Video (Seen Split)}} \\
  \midrule
\textsc{Baseline} \single\ &19.6&20.9&27.5&24.8& 25.8 & 22.9 & 22.1 & 4.0 & 1.0 & 1.2 & 9.8 & 10.6 & 10.7\\
\textsc{GePSAn} \single\ &25.6&-&30.8&-& 28.9 & 26.8 & - & 5.8 & 2.2 & - & 11.8 & 13.4 & -\\
\midrule
\textsc{Baseline} \multiple\ &32.2&-&34.2&-& 35.0 & 30.9 & - & 5.9 & 1.5 & - & 13.7 & 14.8 & -\\
\textsc{GePSAn} \multiple\  &\textbf{36.7}&-&\textbf{38.4}&-& \textbf{37.1} & \textbf{35.0} & - & \textbf{9.3} & \textbf{3.9} & - & \textbf{15.7} & \textbf{17.7} & -\\
 \midrule
 \multicolumn{14}{c}{\textbf{Recipe1M+}} \\
 \midrule
\textsc{Baseline} \single\ &27.0&33.5&29.4&26.7& 24.1 & 22.1 & 22.8 & 7.8 & 4.1 & 4.4 & 11.3 & 13.4 & 13.7\\
\textsc{GePSAn} \single\ &27.2&-&28.5&-& 25.9 & 21.1 & - & 7.5 & 3.4 & - & 11.2 & 12.3 & - \\
\midrule
\textsc{Baseline} \multiple\ &34.7&-&34.6&-& 31.7 & 28.5 & - & 9.4 & 4.9 & - & 14.2 & 16.7 & -\\
\textsc{GePSAn} \multiple\  &\textbf{37.2}&-&\textbf{36.2}&-& \textbf{32.2} & \textbf{29.0} & - &  \textbf{10.7} & \textbf{5.6} & - & \textbf{14.6} &  \textbf{16.9} & - \\
\bottomrule
\end{tabular}
}
\caption{We reproduce and compare the various results corresponding to Tables \ref{res:1} and \ref{res:recipe1m} computed using micro-averaging (\textit{Micro})~\cite{bleu} vs macro-averaging (\textit{Macro})~\cite{sener.etal.2022} of the metrics. When available, we also present the exact numbers reported in the original paper (\textit{Macro}~\cite{sener.etal.2022}). Note, Sener \etal~\cite{sener.etal.2022} report results computed using macro-averaging (\textit{Macro}) only.}
\label{res:recipe1m_reconc}
\end{table*}

\subsection{Results on the YouCookII standard splits}
In Tables~\ref{res:1} and~\ref{res:youcookii_textual} (main paper), we report results for the YouCookII splits proposed by~\cite{sener.etal.2022}, where each split represents a different set of dishes out of the 89 dishes (no overlapping dishes between the different splits). This allowed for comparing the setups where the model has never seen a specific dish before (Unseen Split) vs the setup where the model has seen the dish prepared using other (different) videos (Seen Split). The results were obtained by applying cross-validation on each of the four splits. Here, in Table~\ref{res:recipe1m_stdsplit}, we report results on the original training/validation splits of YouCookII, where the videos are randomly chosen without taking into account which dish they belong to, and hence we can see that it mostly resembles the (Seen Split) setup. 
\begin{table*}[h!]
\centering
\resizebox{0.95\textwidth}{!}{%
\begin{tabular}
{l|| cc | cc | ccc | ccc | ccc}
\toprule
 \multirow{3}{*}{\textbf{Model}} & \multicolumn{2}{c}{ING} & \multicolumn{2}{c}{VERB} & \multicolumn{3}{c}{\textsc{B1}} & \multicolumn{3}{c}{\textsc{B4}} & \multicolumn{3}{c}{\textsc{MET}} \\ 
 \cmidrule(lr){2-14}
  & \small{\textit{Micro}} & \small{\textit{Micro}} & \small{\textit{Micro}} & \small{\textit{Micro}} & \small{\textit{Micro}} & \small{\textit{Macro}} & \small{\textit{Macro}} & \small{\textit{Micro}} & \small{\textit{Macro}} & \small{\textit{Macro}} & \small{\textit{Micro}} & \small{\textit{Macro}} & \small{\textit{Macro}} \\
  &&\small{\cite{sener.etal.2022}}&&\small{\cite{sener.etal.2022}}&&&\small{\cite{sener.etal.2022}}&&&\small{\cite{sener.etal.2022}}&&&\small{\cite{sener.etal.2022}}\\
\midrule
\textsc{Baseline} \single\ & 19.7 & 21.4 & 27.3 & 27.6 & 26.2 & 23.0 & 23.7 & 3.9 & 1.1 & 1.7 & 9.9 & 10.9 & 11.5 \\
\textsc{GePSAn} \single\ & 25.7 &-& 32.2 &-& 30.0 & 27.3 &-& 6.4 & 2.3 &-& 12.2 & 13.8 &-\\
\midrule
\textsc{Baseline} \multiple\ & 33.7 &-& 34.1 &-& 36.2 & 31.5 &-& 6.4 & 1.7 &-& 14.1 & 15.2 &-\\
\textsc{GePSAn} \multiple\ & 37.2 &-& 40.5 &-& 38.4 & 35.8 &-& 9.8 & 4.0 &-& 16.3 &  18.3 &-\\
\bottomrule
\end{tabular}
}
\caption{Future anticipation results on YouCookII Video using the original train/val splits \cite{YC2}. These are the validation results obtained by finetuning different models on the original training split of the YouCookII dataset.}
\label{res:recipe1m_stdsplit}
\end{table*}